\documentclass[runningheads]{llncs}
\usepackage{graphicx}
\usepackage{amsmath}
\usepackage{cite}
\usepackage{algorithmic}
\usepackage{algorithm}
\usepackage{array}
\usepackage{textcomp}
\usepackage{stfloats}
\usepackage{url}
\usepackage{verbatim}
\usepackage{color}
\usepackage{booktabs}
\usepackage{xcolor}
\usepackage{multirow}
\usepackage{enumitem}
\definecolor{citecolor}{HTML}{0071bc} 
\usepackage[colorlinks, linkcolor=red,  anchorcolor=blue, citecolor=citecolor]{hyperref}
\usepackage{float}
\definecolor{SeaGreen4}{RGB}{0,205,102} 
\definecolor{SlateBlue}{RGB}{106,90,205} 
\definecolor{DarkRed}{RGB}{178,34,34} 
\usepackage{times}
\usepackage{soul}
\usepackage[switch]{lineno}
\usepackage{amssymb}
\usepackage{pifont}
\usepackage{makecell}
\usepackage{colortbl}
\definecolor{mygray}{gray}{.9}
\definecolor{mypink}{rgb}{.99,.91,.95}
\definecolor{mycyan}{cmyk}{.3,0,0,0}
\usepackage[misc]{ifsym}
\usepackage{subcaption}  

\begin{document}

\title{ Mamba-FETrack: Frame-Event Tracking via State Space Model } 

 \author{
 Ju Huang\inst{1} \and 
 Shiao Wang\inst{1} \and 
 Shuai Wang\inst{1} \and 
 Zhe Wu\inst{2} \and 
 Xiao Wang (\Letter)\inst{1}* \and 
 Bo Jiang\inst{1} 
\authorrunning{Ju Huang et al.}
%
\institute{1. School of Computer Science and Technology, Anhui University, Hefei 230601, China \\  
2. Pengcheng Laboratory, Shenzhen,  China \\ 
* Corresponding author: Xiao Wang (\email{xiaowang@ahu.edu.cn})
}}



%
\maketitle              

\begin{abstract}
RGB-Event based tracking is an emerging research topic, focusing on how to effectively integrate heterogeneous multi-modal data (synchronized exposure video frames and asynchronous pulse Event stream). Existing works typically employ Transformer based networks to handle these modalities and achieve decent accuracy through input-level or feature-level fusion on multiple datasets. However, these trackers require significant memory consumption and computational complexity due to the use of self-attention mechanism. This paper proposes a novel RGB-Event tracking framework, Mamba-FETrack, based on the State Space Model (SSM) to achieve high-performance tracking while effectively reducing computational costs and realizing more efficient tracking. Specifically, we adopt two modality-specific Mamba backbone networks to extract the features of RGB frames and Event streams. Then, we also propose to boost the interactive learning between the RGB and Event features using the Mamba network. The fused features will be fed into the tracking head for target object localization. Extensive experiments on FELT and FE108 datasets fully validated the efficiency and effectiveness of our proposed tracker. Specifically, our Mamba-based tracker achieves 43.5/55.6 on the SR/PR metric, while the ViT-S based tracker (OSTrack) obtains 40.0/50.9. The GPU memory cost of ours and ViT-S based tracker is 13.98GB and 15.44GB, which decreased about $9.5\%$. The FLOPs and parameters of ours/ViT-S based OSTrack are 59GB/1076GB and 7MB/60MB, which decreased about $94.5\%$ and $88.3\%$, respectively. We hope this work can bring some new insights to the tracking field and greatly promote the application of the Mamba architecture in tracking. 
The source code of this work will be released on \url{https://github.com/Event-AHU/Mamba_FETrack}. 
\end{abstract}

\section{Introduction}

\begin{figure}
\centering
\includegraphics[width=0.75\linewidth]{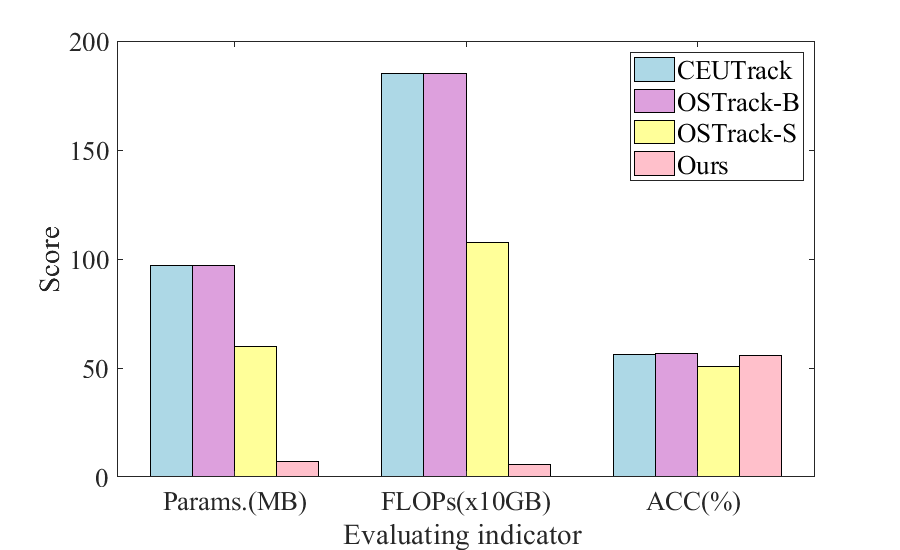}
\caption{Comparison of the parameters, FLOPs, and accuracy of our proposed Mamba-FETrack and other strong RGB-Event trackers.} 
\label{fig:ComparisonParams}
\end{figure}

Visual tracking is a widely exploited research topic in computer vision that aims to predict the locations of the target object in the subsequent video frames given its initial localization. It has been widely used in many practical scenarios, including smart video surveillance, UAVs, autonomous vehicles, military fields, etc. Current tracking algorithms are usually developed based on RGB frames captured using the visible camera and the overall tracking performance has been boosted by deep learning techniques greatly in recent years. These deep learning based trackers can be divided into four main streams, including proposal classification based tracking (e.g., MDNet~\cite{Nam2015LearningMC}, RT-MDNet~\cite{jung2018real}), Siamese network based tracking (e.g., SiamFC~\cite{bertinetto2021fullyconvolutional}, SiamRPN++~\cite{xu2020siamfc++}, SINT~\cite{Tao2016SiameseIS}, SINT++~\cite{Wang2018SINTRV}), Transformer network based tracking (e.g., TransT~\cite{Chen2021TransformerT}, STARK~\cite{Yan2021LearningST}, MixFormer~\cite{Cui2022MixFormerET}, AiATrack~\cite{gao2022aiatrack} ). There are also some works that attempt to address the issues caused by the local search strategy used in most tracking frameworks, such as TANet~\cite{liu2020tanet}, DeepMTA~\cite{yang2020interpretable}, and GlobalTrack~\cite{huang2020globaltrack}. Although the trackers already achieved good performance on existing benchmark datasets, the tracking results under extremely challenging scenarios (e.g., fast motion and low illumination) are still far from satisfying.

To further improve the tracking performance in challenging scenarios, this paper resorts to the newly developed Event cameras (e.g., DVS346, Prophesee, CeleX-V) to provide the external cues. Event camera can capture motion cues well due to their unique imaging principle. To be specific, the Event camera asynchronously records the pixels and it records the pixel only when its variation of light intensity exceeds a given threshold. Event camera also performs better than the visible camera on high dynamic range, low energy consumption, low data storage, etc~\cite{Gallego2019EventBasedVA}. However, the Event camera captures nothing or very few Event points when the target object is motionless. It also fails to record the color or detailed texture information of the tracked target object. Thanks to the good results provided by the visible camera in these cases, we can combine the two cameras to achieve robust and high-performance tracking.

Based on this idea, some RGB-Event visual trackers have already been proposed in recent years. Specifically, the CMT~\cite{Wang2021VisEventRO} proposed by Wang et al. introduces a new cross-modality Transformer module for the fusion of RGB and Event data. Tang et al.~\cite{tang2022coesot} propose a unified adapter-based Transformer network as their backbone which achieves feature extraction, fusion, and tracking simultaneously. Zhang et al.~\cite{Zhang2022SpikingTF} propose a hybrid SNN-Transformer framework for energy-efficient visual tracking. Although these works achieve good results on existing benchmark datasets, however, these tracking algorithms are still limited by the following issues: 
\textbf{1).} These trackers are all developed based on the Transformer network whose computational complexity is very high, as the complexity of the attention computation is $\mathcal{O}(N^2)$. 
\textbf{2).} The memory requirement on the GPU is very high in the training and tracking phase, as the KV cache needs to be stored for subsequent processing. 
Obviously, the tracking speed of these trackers is limited due to the aforementioned issues. 
Thus, it is natural to raise the following question: \textit{How can we design a new RGB-Event tracking framework that achieves a good balance between high-performance and low model complexity?}

Recently, the state space model (SSM) has drawn more and more attention in the artificial intelligence community, especially in natural language processing and computer vision~\cite{wang2024SSMSurvey}. The SSM is an attention-free model with the complexity $\mathcal{O}(N)$. On the other hand, current SSM based vision models~\cite{Nguyen2022S4NDMI, Smith2022SimplifiedSS, Liu2024VMambaVS,zhu2024vision} also demonstrate that they can achieve similar or even higher performance on large-scale classification tasks, such as ImageNet-1K~\cite{Deng2009ImageNetAL}. The FLOPs and memory requirement are also decreased significantly compared with the Transformer based networks. It has been widely used in many domains and tasks, such as RGB/Video-based recognition~\cite{Li2024MambaNDSS}, segmentation~\cite{Xing2024SegMambaLS,Ma2024UMambaEL,Ruan2024VMUNetVM}, graph-based modeling~\cite{tang2023modeling,Wang2024GraphMambaTL,Behrouz2024GraphMT}, point cloud~\cite{Liang2024PointMambaAS,Zhang2024PointCM,Liu2024PointMA}, Event stream~\cite{Zubic2024StateSM}, time series data~\cite{islam2022long,Wang2023SelectiveSS}, etc. However, the performance of the state space model based RGB-Event tracking task is still unknown. In this paper, we propose a novel RGB-Event tracking framework based on the State Space Model (SSM) to replace the Transformer network. For the details of our framework, it takes the RGB frames and Event stream as the input, then, two modality-specific Mamba backbone networks are designed to encode the RGB and Event inputs. Also, we introduce the Mamba to boost the interactive learning between dual modalities. The fused features will be fed into the tracking head for target object localization. As shown in Fig.~\ref{fig:ComparisonParams}, our Mamba-FETrack achieves much lower FLOPs and parameters, meanwhile, maintaining comparable or even better results than the ViT-B and ViT-S based OSTrack RGB-Event tracker. 
An overview of our proposed Mamba based RGB-Event tracking framework is illustrated in Fig.~\ref{fig:framework}.

To sum up, we draw the contributions of this paper as the following three aspects: 

1). We propose the \textit{first} state space model based RGB-Event tracking framework which achieves a good tradeoff between high performance and low complexity, termed Mamba-FETrack. It improves the GPU memory cost, FLOPs, and parameters of ViT-S based OSTrack (RGB-Event version) by $9.5\%$, $94.5\%$ and $88.3\%$, respectively, meanwhile, exceeds this tracker on tracking performance. 

2). We introduce a simple but effective \textit{FusionMamba block} which boosts the interactive feature learning between the RGB and Event features effectively. 

3). We conduct extensive experiments on two large-scale RGB-Event tracking datasets (i.e., FELT~\cite{wang2024longterm}, and FE108~\cite{Zhang2021ObjectTB}) which fully validated the effectiveness of our proposed Mamba based RGB-Event tracker.

\section{Related Work} 

The dual-modal tracking model, integrating RGB and Event data, primarily relies on Transformer architecture. Our proposed Mamba-FETrack utilizes a streamlined Mamba backbone to achieve efficiency within the model. In this section, we examine current mainstream Frame-Event Tracking algorithms and State Space models (SSM). More related works can be found on the following paper list~\footnote{\url{github.com/Event-AHU/Mamba_State_Space_Model_Paper_List}}.

\subsection{Frame-Event Tracking} 
\hspace{0.5cm}
The Event camera offers benefits such as high dynamic range and minimal delay, while the RGB camera captures detailed texture information of the scene. Considering the complementary advantages of both modalities, an increasing number of researchers are leveraging data from both modalities concurrently to attain precise and robust visual tracking.
Zhang et al.~\cite{Zhang2021ObjectTB} effectively fuse visual cues from both Frame-Event data by employing cross-domain attention schemes and a specially crafted weighting mechanism, thereby enhancing the performance of single target tracking.
CMT~\cite{Wang2021VisEventRO}, proposed by Wang et al., employs cross-modality transformer to facilitate information interaction between Event images and visible images.
AFNet~\cite{Zhang2023FrameEventAA}, introduced by Zhang et al., presents a framework for high frame rate tracking. It utilizes multi-modal alignment and fusion module to effectively integrate meaningful information from two modalities at different measurement rates.
Tang et al. propose a unified tracking framework, CEUTrack~\cite{tang2022coesot}, based on RGB frames and Color-Event voxels. The framework employs a single-stage backbone network to simultaneously accomplish feature extraction, fusion, matching, and interactive learning.
Zhu et al.~\cite{Zhu2023CrossmodalOH} propose a mask modeling strategy. This involves randomly masking tokens of modalities to encourage the Vision Transformer (ViT) to bridge the distribution gap between the two modalities, thereby enhancing the model's capability.
ODTrack~\cite{Zheng2024ODTrackOD}, proposed by Zheng et al., extensively correlates context between video frames via online token propagation to establish Event correlations between image pairs.
The majority of bi-modal approaches leveraging RGB and Event data are grounded in Transformer architecture. However, unlike our proposed Mamba-FETrack, which employs a simpler backbone (Mamba) to achieve efficiency within the model.

\subsection{State Space Model and Mamba} 

\hspace{0.5cm}
The State Space Model (SSM) originally served as a mathematical framework for characterizing the dynamics of dynamic systems.
Gu et al.~\cite{Gu2021CombiningRC} introduce an innovative model termed the linear State Space Layer (LSSL). This model integrates the strengths of recurrent neural networks (RNNs), temporal convolution, and neural differential equations (NDEs), effectively mitigating the limitations in model capacity and computational efficiency.
To address the issue of gradient disappearance/explosion encountered by SSMS when modeling longer sequences, the HiPPO model~\cite{gu2020hippo} combines recursive memory and optimal polynomial projection concepts. This approach notably enhances the performance of recursive memory and aids SSMS in handling long sequences and long-term dependencies more effectively.
Building upon this foundation, Gu et al. introduced the Structured State Space Sequence Model (S4)~\cite{gu2021efficiently}. This novel parameterization method, derived from the basic State Space Model (SSM), addresses long-distance dependencies through both mathematical and empirical approaches, significantly enhancing computational efficiency.

In recent years, there has been a growing emphasis on the State Space Model (SSM), originally applied to natural language processing tasks. Mamba~\cite{Gu2023MambaLS} proposed a time-varying state-space model based on a selection mechanism, effectively modeling long sequences. Following Mamba's success, researchers have applied it to various research fields. In vision tasks, Vim~\cite{zhu2024vision} introduces a generic vision backbone with bidirectional Mamba blocks, marking image sequences with position embeddings, and compressing visual representations with bidirectional state space models. VMamba~\cite{Liu2024VMambaVS} introduces the Cross-Scan Module (CSM) to traverse spatial domains and convert non-causal visual images into ordered patch sequences. In the realm of multimodal and multimedia, S4ND~\cite{Nguyen2022S4NDMI} extends SSM's continuous signal modeling capabilities to multidimensional data, including images and video. Pan-Mamba~\cite{He2024PanMambaEP} employs channel swapping Mamba and cross-modal Mamba to achieve efficient cross-modal information exchange and fusion. 
Wang et al. summarize existing SSM-based models and related applications in the survey~\cite{wang2024SSMSurvey}. 
While current bi-modal approaches, relying on RGB and Event data, are mainly based on Transformer architecture, our efforts have successfully applied Mamba in this context.

\section{Preliminary: SSMs and Mamba} 
The State Space Model (SSM) is derived from the classical Kalman filter~\cite{kalman1960new}. It maps the input sequence $x(t) \in \mathbb{R}^{L}$ into output $y(t) \in \mathbb{R}^{L}$ via hidden state $h(t) \in \mathbb{R}^{N}$. In the continuous state, the specific expression of SSM is described by a set of ordinary differential equations:
\begin{equation}  
\begin{aligned} 
\label{continuous SSMs}
h'(t) = \mathbf{A}h(t) + \mathbf{B}x(t),    \\  
y(t) = \mathbf{C}h(t) +  \mathbf{D}x(t).
\end{aligned}  
\end{equation}
where $\mathbf{A} \in \mathbb{R}^{N \times N} $ is the state matrix, $\mathbf{B} \in \mathbb{R}^{N \times L} $ is the input matrix , $\mathbf{C} \in \mathbb{R}^{L \times N} $ is the output matrix, $\mathrm{~and~} \mathbf{D} \in \mathbb{R}^{L \times L}$ is the feed-through matrix. $h'(t) \in \mathbb{R}^{N}$ is the derivative of the hidden state.

By introducing the time scale parameter $\Delta$, the zero-order hold rule is adopted to discretize the continuous state space equation of SSM:
\begin{equation}  
\begin{aligned} 
\label{discretize SSMs}
{h}_k &=  \bar{\mathbf{A}}{h}_{k-1} + \bar{\mathbf{B}}{x}_k, \\
{y}_k &= \bar{\mathbf{C}}{h}_k, 
\end{aligned}  
\end{equation}
where $h_{k-1}$ and $h_k$ represent the discrete hidden state, $x_k$ and $y_k$ are the discrete inputs and outputs. $\bar{\mathbf{A}}$, $\bar{\mathbf{B}}$, and  $\bar{\mathbf{C}}$  are discrete parameters of the system, and the process of discretization through time scale parameter $\Delta$ can be expressed as:
\begin{equation}  
\begin{aligned} 
\label{discretize parameter}
\overline{\mathbf{A}} &= \exp(\Delta\mathbf{A}), \\
\overline{\mathbf{B}} &= (\Delta\mathbf{A})^{-1}(\exp(\Delta\mathbf{A}) - \mathbf{I}) \cdot \Delta\mathbf{B},\\
\overline{\mathbf{C}} &= \mathbf{C}.
\end{aligned}  
\end{equation}
More details about the working principle of Mamba can be found in~\cite{Gu2023MambaLS, wang2024SSMSurvey}.

\section{Our Proposed Approach}

\subsection{Overview}

In this section, we introduce the overall framework of the newly proposed method Frame-Event Tracking using State Space Model (Mamba-FETrack). As illustrated in Fig.~\ref{fig:framework}, our framework takes the template patch and search patch of the RGB image as well as the Event image as inputs first. Subsequently, it obtains the corresponding token representations through the linear projection layer and position encoding. RGB Mamba blocks and Event Mamba blocks are utilized to extract features from the RGB image and Event image, respectively. Next, a FusionMamba block is employed to model the interaction and relationship between RGB modal and Event modal. Finally, the fused features are fed into the tracking head to predict the locations of target object.

\subsection{Network Architecture}

\begin{figure*}
\centering
\includegraphics[width=\textwidth]{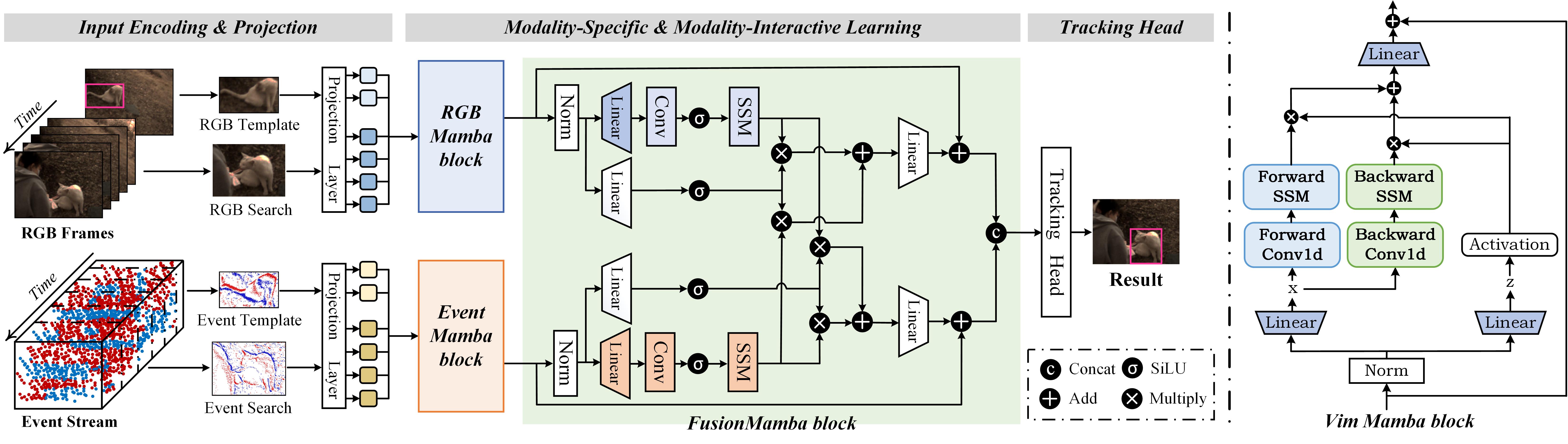}
\caption{An overview of our proposed Frame-Event tracking via state space model.}
\label{fig:framework}
\end{figure*}

\noindent 
\textbf{Input Representation.} 
The RGB frames and Event stream are both adopted as the input of our Mamba-FETrack framework. Here, RGB frames are denoted as $I=\{I_1, I_2,\\ \ldots, I_N\}$, where $I_i$, $i \in [1, N]$, represents the $i$-th RGB frame and $N$ represents the total number of frames in a video. The Event stream is defined as $\mathcal{E} = \{e_{1}, e_{2}, \ldots, e_{M}\}$, where $e_j$, $j \in [1, M]$, represents the $j$-th Event point and $M$ represents the total number of Event points in a video. For RGB frames $I$, we use the standard data processing in OSTrack~\cite{ye2022joint} to get RGB template patch $Z_I$ and RGB search patch $X_I$. For Event stream $\mathcal{E}$, we stack Event points into Event images aligned with RGB frames according to the camera's exposure time and use a similar data processing method to get Event template patch $Z_E$ and Event search patch $X_E$. Then, the patch sequence is projected into patch embeddings through the linear projection layer, and the learnable position embedding is added to the patch embeddings of the template and search respectively. The corresponding token representations are finally obtained: $H_{\text{I}}^z\in \mathbb{R}^{B \times N_1 \times C}, H_{\text{I}}^x\in \mathbb{R}^{B \times N_2 \times C} $ \text{ and } $H_{\text{E}}^z\in \mathbb{R}^{B \times N_1 \times C}, H_{\text{E}}^x\in \mathbb{R}^{B \times N_2 \times C} $, where B and C are the batch size and channel dimension, $N_1$ and $N_2$ are the number of tokens of template and search.

\noindent 
\textbf{Modality-Specific Mamba Backbone.} 
Inspired by vision Mamba~\cite{zhu2024vision}, we utilize RGB and Event Mamba blocks to extract features from the RGB images and Event images, respectively, thereby obtaining specific features of the two modalities. For the feature extraction of RGB images, we concatenate the RGB template tokens and the RGB search tokens based on the number of tokens to obtain the input token sequence $H_{\text{I}}^0$ =[$H_{\text{I}}^z, H_{\text{I}}^x$]. Subsequently, we utilize a series of stacked Vision Mamba blocks to extract specific features of RGB, denoted as \(F_{\mathrm{rgb}}\). Fig.~\ref{fig:framework} (right) shows the architecture of the Modality-Specific Mamba Backbone. 

The detailed process of the Mamba block is outlined as follows: First, $H_{\text{I}}^{l-1}$ is normalized through a normalization layer. Then, $z$ and $x$ are obtained via two linear projection layers:
\begin{equation}
\begin{aligned}
\label{LayerNorm}
z &= (Linear^{z}({Norm}(H_{I}^{l-1}))), \quad & x &= (Linear^{x}({Norm}(H_{I}^{l-1}))) \\
\end{aligned}
\end{equation}
Next, we process $x$ in both forward and backward directions. For each direction, the 1D convolution and SiLU activation function are applied to $x$ to produce \(x^{\prime}\):
\begin{equation}
\begin{aligned}
\label{x'}
{x}^{\prime} = {SiLU}({Conv1d}({x}))
\end{aligned}
\end{equation}
where $Conv1d$ denotes the depth-wise convolution, $SiLU$ denotes the SiLU activation function. Furthermore, \(x^{\prime}\) is linearly projected to yield the input matrix \textbf{B}, output matrix \textbf{C}, and the time scale parameter \textbf{$\Delta$}. Leveraging \textbf{$\Delta$}, we compute the discrete matrices $\bar{\textbf{A}}$ and $\bar{\textbf{B}}$. These operations can be expressed as:
\begin{equation}
\begin{gathered}
\begin{aligned}
\label{ABC}
\textbf{B}, \textbf{C}, \textbf{$\Delta$} &= {Linear}(x^{\prime}), \quad 
\bar{\textbf{A}},\bar{\textbf{B}}={ZOH}(\textbf{A},\textbf{B},\textbf{{$\Delta$}})
\end{aligned}
\end{gathered}  
\end{equation}
where $ZOH$ denotes the zero-order hold rule. Subsequently, $y_{forward}$ and $y_{backward}$ are calculated using SSM. Then, $y_{forward}$ and $y_{backward}$ are gated by $z$ and added together to obtain \(y^{\prime}\). The formulas are defined below:
\begin{equation}  
\begin{gathered}
\begin{aligned}
\label{forward-backward}
y^{\prime} &= y_{\text{forward}} \odot {SiLU}(z) + y_{\text{backward}} \odot {SiLU}(z), \\ 
y_{P} &= \mathrm{SSM}_{P}(x^{\prime}), \quad P \in \{ \text{forward}, \text{backward} \}
\end{aligned}
\end{gathered}  
\end{equation}
Specifically, the calculation process of SSM can be expressed as:
\begin{equation}  
\begin{gathered}
\begin{aligned}
\label{SSM}
h^{\prime}=\bar{\textbf{A}}h+\bar{\textbf{B}}x^{\prime},\quad y=\textbf{C}h^{\prime} \\
\end{aligned}
\end{gathered}  
\end{equation}
Finally, the input token sequence $H_ {\text {I}} ^ {l-1}$ is also considered to produce the final output feature $H_{\text{I}}^l$ via residual connection, where $l \in [1, L]$ and \textit{L} represents the total number of Mamba blocks. For the feature extraction of Event images, we apply the same method as above to obtain the specific feature of the Event, denoted as \(F_{\mathrm{event}}\).

\noindent 
\textbf{Modality-fusion Mamba Network. } 
Inspired by the Cross-Modal Mamba block in Pan-Mamba~\cite{He2024PanMambaEP}, we employ modality interactive learning to enhance feature interaction and fusion between RGB and Event modalities. The detailed process of learning Event information for 
\(F_{\mathrm{rgb}}\) is as follows: First, \(x_{rgb}\) and \(x_{event}\) are derived from \(F_{\mathrm{rgb}}\) and \(F_{\mathrm{event}}\) through the application of the normalized layer and the linear projection layer, respectively. Concurrently,  \(F_{\mathrm{rgb}}\) is processed by a linear projection layer to obtain $z_{rgb}$. The specific formulas are as follow:
\begin{equation}
\begin{aligned}
\label{x_rgb X_event}
x_{m} &= (Linear_{m}^{x}({Norm_{m}}(F_{{m}}))), \quad m \in \{ {rgb}, {event}\}\\
\end{aligned}
\end{equation}
\begin{equation}
\begin{aligned}
\label{LayerNorm}
z_{rgb} &= (Linear_{rgb}^{z}({Norm_{rgb}}(F_{{rgb}}))) \\
\end{aligned}
\end{equation}
Then, \(y_{rgb}\) and \(y_{event}\) are computed using one-dimensional convolution with the SiLU activation function and SSM:
\begin{equation}
\begin{aligned}
\label{linear}
y_{m} &= \mathrm{SSM}({SiLU}({Conv1d}({x_{m}}))), \quad m \in \{ {rgb}, {event}\}\\
\end{aligned}
\end{equation}
Then, \(y_{rgb}\) and \(y_{event}\) are gated by $z_{rgb}$,  refining the outputs of \(y_{rgb}^{\prime}\) and \(y_{event}^{\prime}\):
\begin{equation}
\begin{aligned}
\label{LayerNorm}
y_{m}^{\prime} &= y_{m} \odot {SiLU}(z_{rgb}), \quad m \in \{ {rgb}, {event}\} \\
\end{aligned}
\end{equation}
Finally, \(y_{\text{rgb}}^{\prime}\) and \(y_{\text{event}}^{\prime}\) are added, and then residual on \(F_{\mathrm{rgb}}\) to get the RGB fusion feature \(\tilde{F}_{\text{rgb}}\). The process of obtaining \(\tilde{F}_{\text{event}}\) and the fusion feature of the event, follows the same steps as described above. Through modality interactive learning, the comprehension of complementary modalities is heightened.

\subsection{Tracking Head and Loss Function} 

The tracking head adopts the same approach as OSTrack~\cite{ye2022joint}. We concatenate the features of the search from \(\tilde{F}_{\text{event}}\) and \(\tilde{F}_{\text{rgb}}\) along the channel dimension, and then feed them into the tracking head to determine the location of the target object.
Firstly, the input features are transformed into a feature map, and then fed into a series of stacked Convolution-Batch Normalization-ReLU (Conv-BN-ReLU) layers to generate the outputs. The outputs comprise three main components: the target classification score map, the local offset, and the normalized size of the bounding box. Finally, the ultimate target bounding box is computed based on the output.

We employ three distinct loss functions during the training like OSTrack~\cite{ye2022joint}, which encompass the focal loss, L1 loss, and GIoU loss, to ensure comprehensive optimization. The loss function can be formulated as:
\begin{equation}
\mathcal{L} = \lambda_1 \mathcal{L}_{\text{focal}} + \lambda_2 \mathcal{L}_1 + \lambda_3 \mathcal{L}_{\text{GIoU}}
\end{equation}
where \(\lambda_{1}=1\), \(\lambda_{2}=14\), and \(\lambda_{3}=1\) are the weight coefficients in our experiment.

\section{Experiment} 

\subsection{Dataset and Evaluation Metric}
In the experiments, we validate our tracker on two large-scale benchmark datasets, i.e., \textbf{FE108} and \textbf{FELT} dataset. A brief introduction to the two datasets are given below. 

\noindent
$\bullet$ \textbf{FE108}: The dataset is captured using a grayscale DVS346 Event camera and comprises 76 training videos and 32 testing videos. It includes over 1132K annotations across more than 143K images and their corresponding Events. The dataset encompasses various degraded conditions for tracking, such as motion blur and high dynamic range, enhancing its versatility for research and evaluation purposes.

\noindent
$\bullet$ \textbf{FELT}: The FELT dataset, collected by the DVS346 Event camera, offers a unique blend of RGB frames (346×260) and Event stream. This expansive dataset comprises 742 sequences, meticulously divided into 520 for training and 222 for testing. Each sequence spans over a minute and encompasses a grand total of 1,594,474 frames across 45 distinct categories. Additionally, FELT features a rich annotation of 14 challenge attributes, enhancing its depth and utility for diverse tracking scenarios.

For the evaluation metrics, we adopted the three widely used evaluation metrics in tracking tasks: \textbf{Precision Rate (PR)}, \textbf{Normalized Precision Rate (NPR)}, and \textbf{Success Rate (SR)}.

\subsection{Implementation Details} 
Our proposed Mamba-FETrack framework is extended based on OSTrack~\cite{ye2022joint} which adopts the vision Transformer network as its backbone, while ours is the lightweight vision Mamba~\cite{zhu2024vision}. We set the learning rate, weight decay, and batch size to 0.0004, 0.0001, and 32, respectively. AdamW~\cite{loshchilov2018adamw} is chosen as the optimizer. Our implementation is developed in Python, utilizing the PyTorch~\cite{paszke2019pytorch} framework, and the experiments are conducted on a server equipped with an Intel (R) Xeon (R) Gold 5318Y CPU@2.10GHz and RTX 3090 GPU.

\subsection{Comparison with Other SOTA Algorithms}  

\begin{table}
\center
\small     
\caption{Experimental results (SR/PR) on FE108 dataset.} 
\label{FE108table}
\begin{tabular}{c|cccccc}
\hline  \toprule [0.5 pt] 
\textbf{Tracker}  &\textbf{SiamRPN}  &\textbf{SiamBAN}  &\textbf{SiamFC++}   &\textbf{KYS}  &\textbf{CLNet}  &\textbf{CMT-MDNet} \\
\textbf{} &\cite{Li2018HighPV}  &\cite{Chen2020SiameseBA}  &\cite{xu2020siamfc++}  &\cite{bhat2020know}  &\cite{dong2020clnet}  &\cite{Wang2021VisEventRO} \\
\hline
\textbf{AUC/PR}  &21.8/33.5 &22.5/37.4 &23.8/39.1 &26.6/41.0 &34.4/55.5 &35.1/57.8  \\
\hline
\textbf{Tracker}  &\textbf{ATOM}     &\textbf{DiMP}     &\textbf{PrDiMP}     &\textbf{CMT-ATOM}   &\textbf{CEUTrack}       &\textbf{Ours} \\ 
\textbf{} &\cite{Danelljan2018ATOMAT}  &\cite{Bhat2019LearningDM}  &\cite{Danelljan2020ProbabilisticRF}  &\cite{Wang2021VisEventRO}  &\cite{tang2022coesot}  & -  \\
\hline
\textbf{AUC/PR}  &46.5/71.3 &52.6/79.1 &53.0/80.5 &54.3/79.4 &55.58/84.46  & 58.71/90.95 \\
\hline  \toprule [0.5 pt]  
\end{tabular}
\end{table}	

\noindent
\textbf{Results on FE108 dataset. }
As shown in Table~\ref{FE108table}, we compared the experimental results of our proposed Mamba-FETrack and other tracking methods on the FE108 dataset. The experimental results show that our baseline method CEUTrack~\cite{tang2022coesot} achieves 55.58/84.46 on SR/PR, and at the same time, our method surpasses the baseline to achieve 58.71/90.95. Compared to other comparative trackers (e.g., DiMP\cite{Bhat2019LearningDM} and PrDiMP\cite{Danelljan2020ProbabilisticRF}), our method shows a greater advantage, with a precision rate 11.85 percentage points higher than DiMP\cite{Bhat2019LearningDM}. These experimental results demonstrate the effectiveness of our method on the FE108 dataset.

\begin{table}
\centering
\small   
\caption{Tracking results on FELT SOT dataset (SR/PR). Note that, OSTrack* indicates that we first perform image fusion at the input level on the two modalities.} 
\label{FELT_benchmark_results}
\setlength{\tabcolsep}{12pt} 
\resizebox{\textwidth}{!}{ 
\begin{tabular}{l|c|ccccc}
\hline \toprule [0.5 pt]  
\textbf{Trackers} & \textbf{Publish}   & \textbf{RGB}  &\textbf{Event}   &\textbf{RGB+Event}    &\textbf{FPS}  \\
\hline
\textbf{01. STARK  }  \cite{Yan2021LearningST}      &ICCV21            &45.6/58.2        &39.3/50.8      &45.7/59.4        & 42           \\ 
\hline
\textbf{02. GRM  } \cite{Gao2023GeneralizedRM}      &CVPR23            & 44.7/55.9       &39.2/48.9    &44.5/55.9        & 45           \\ 
\hline
\textbf{03. PrDiMP  } \cite{Danelljan2020ProbabilisticRF}      &CVPR20            &43.8/54.7        &34.9/44.5      &43.8/55.2       & 30           \\ 
\hline
\textbf{04. DiMP } \cite{Bhat2019LearningDM}      &ICCV19            &43.3/54.5        &37.8/48.5      &43.5/55.1        &   43         \\ 
\hline
\textbf{05. SuperDiMP } \cite{Bhat2019LearningDM}      &-            &43.0/53.5       &37.8/47.8      &43.0/54.2        &  -          \\ 
\hline
\textbf{06. TransT} \cite{Chen2021TransformerT}      &CVPR21            & 42.2/53.0      &35.2/45.1   & 34.6/44.3       &50            \\ 
\hline
\textbf{07. TOMP50} \cite{Mayer2022TransformingMP}      &CVPR22            &41.2/51.8      &37.7/49.2    &43.4/55.2        & 25           \\ 
\hline
\textbf{08. ATOM}  \cite{Danelljan2018ATOMAT}     &CVPR19            &37.5/47.0       &22.3/28.4     &36.2/45.9       &30            \\ 
\hline
\textbf{09. KYS}  \cite{bhat2020know}     &ECCV20            & 35.9/45.0      &22.5/29.5      &33.1/42.4        & 20           \\ 
\hline
\textbf{10. OSTrcak-B} \cite{ye2022joint}      &ECCV22            & -     & -    &45.0/56.6        &63            \\ 
\hline
\textbf{11. OSTrack-S} \cite{ye2022joint}      & ECCV22           &-      &-      &40.0/50.9        &84            \\ 
\hline
\textbf{12. OSTrack*  } \cite{ye2022joint}      &ECCV22            &45.2/56.3        &37.4/46.9      & 32.5/40.3       & 105           \\ 
\hline
\textbf{13. AFNet} \cite{Zhang2023FrameEventAA}      &CVPR23           &-       & -     &28.9/36.6        &36           \\ 
\hline
\textbf{14. Mamba-FETrack}  &-            &43.3/55.1        &38.3/48.3      &43.5/55.6       &  24           \\ 
\hline  \toprule [0.5 pt] 
\end{tabular}
}
\end{table}

\noindent
\textbf{Results on FELT dataset.} 
As shown in Table~\ref{FELT_benchmark_results}, we also report the experimental results of our tracker and other methods on the long-term RGB-Event SOT dataset FELT. It is obvious that our method achieves excellent accuracy on the FELT dataset, which is 43.5 and 55.6 on SR and PR, respectively. Compared to other tracking methods such as AFNet\cite{Zhang2023FrameEventAA}, our method demonstrates greater advantages in both accuracy and model parameters. Specifically, compared to OSTrack*\cite{ye2022joint}, our method outperforms +11 and +15.3 on SR and PR. These experiments demonstrate that our method also has significant advantages on long sequence datasets like FELT, especially in terms of training efficiency, which far exceeds other methods.

\subsection{Ablation Study}  
\noindent
\textbf{Effectiveness on the Fusion of RGB Frame and Event Stream for Tracking.} 
To demonstrate the effectiveness of our model using RGB frame and Event stream as inputs, we conduct a comparative experiment to analyze the model's performance with different types of data inputs. We present the results obtained from three different inputs: only RGB frame, only Event stream, both RGB frame and Event stream simultaneously. As shown in Table~\ref{input ablation}, when only RGB frame is used as inputs, our method achieves 43.3, 55.1, and 51.7 on SR, PR, and NPR, respectively. When only Event stream is used as input, the tracking performance is poor. Our method achieves 38.3, 48.3, and 45.7 on SR, PR, and NPR, respectively. When both RGB frame and Event stream are used simultaneously, the performance can be improved to 43.5, 55.6, and 51.9 on SR, PR, and NPR, respectively. It's evident that the model effectively integrates the complementary information from both modalities through modality interactive learning, resulting in more accurate target tracking.




\begin{figure}
\resizebox{\textwidth}{!}{%
\begin{minipage}[b]{0.5\linewidth}
\centering
\small
\captionsetup{justification=centering}
\captionof{table}{Ablation study on data inputs (FELT dataset).} 
\label{input ablation}
\begin{tabular}{l|lll}
\hline \toprule [0.5 pt]  
\textbf{Input Data}    &\textbf{SR}   & \textbf{PR}  & \textbf{NPR}  \\
\hline
\text{1. Event Frames }           &38.3   &48.3    &45.7  \\
\text{2. RGB Frames }             &43.3   &55.1    &51.7  \\
\text{3. Event+RGB}             &43.5   &55.6    &51.9  \\
\hline \toprule [0.5 pt]  
\end{tabular}
\end{minipage}%
\begin{minipage}[b]{0.5\linewidth}
\centering
\small
\captionsetup{justification=centering}
\captionof{table}{Ablation study on backbone networks (FELT dataset).} 
\label{backbone ablation}
\begin{tabular}{c|lll}
\hline \toprule [0.5 pt]  
\textbf{Backbone}    &\textbf{SR}   & \textbf{PR}  & \textbf{NPR}  \\
\hline
\text{1. ViT-B }              &45.0     &56.6    &53.9  \\
\text{2. ViT-S }             &40.0     &50.9    &49.1  \\
\text{3. Vim-S}              &43.5   &55.6    &51.9  \\
\hline \toprule [0.5 pt]  
\end{tabular}
\end{minipage}%
}
\end{figure}

\noindent
\textbf{Effectiveness on SSM for Tracking.} 
To validate the effectiveness of the state space model for tracking, we conduct a comparative experiment between the SSM-based network and the ViT-B based network. Our method utilizes the weights of Vim-small~\cite{zhu2024vision} as pre-trained model, fine-tuned on the FELT dataset. Similarly, we use the weights of ViT-S~\cite{Dosovitskiy2020AnII} and ViT-B~\cite{Dosovitskiy2020AnII} as pre-trained models to conduct comparative experiments, thus verifying the effectiveness of SSM. As shown in Table~\ref{backbone ablation}, our method achieves a similar performance to the ViT-B based tracker. Compared with the method that only uses ViT-S as the backbone, our method has a +3.5 and +4.7 higher success rate and precision rate, respectively. 




\begin{figure}
\begin{minipage}[t]{0.5\linewidth}
\centering
\begin{table}[H]
\centering
\small
\caption{Memory Usage of SSM and ViT} 
\label{Memory Usage}
\begin{tabular}{c|ccc}
\hline \toprule [0.5 pt]  
\textbf{Backbone}  &\textbf{Vim-S}  &\textbf{ViT-S}  &\textbf{ViT-B}   \\
\hline
Memory Usage  &14320MB       &15810MB           &23650MB        \\
\hline \toprule [0.5 pt]  
\end{tabular}
\end{table}
\end{minipage}%
\begin{minipage}[t]{0.5\linewidth}
\centering
\begin{table}[H]
\centering
\small
\caption{Hybrid SSM-Transformer}
\label{SSM-ViT}
\begin{tabular}{c|cc}
\hline \toprule [0.5 pt]  
\textbf{Method}  &\textbf{Ours}  &\textbf{SSM-Transformer}   \\
\hline
PR/SR  &55.6/43.5      &55.9/43.8         \\
\hline \toprule [0.5 pt]  
\end{tabular}
\end{table}
\end{minipage}%
\end{figure}

\noindent
\textbf{Effectiveness on Decreasing Memory Usage of SSM.} 
One of the advantages of the state space model over Transformer is its linear complexity. To verify whether the state space model can significantly reduce GPU usage, we compare the GPU memory of our method and ViT based method under the same settings. During training, we set the batch size to 32. As shown in Table~\ref{Memory Usage}, our method uses similar GPUs as ViT-S, but ViT-B uses 23650M GPU memory. It can be found that our method can achieve the same performance as Transformer even without self-attention, while significantly improving the efficiency of computing and storage.

\noindent
\textbf{Effectiveness on Hybrid SSM-Transformer for Frame-Event Tracking.} 
To ensure the accuracy of our model, we added some Transformer blocks to improve the robustness of the model. Specifically, we added a layer of Transformer block after the backbone of each branch. At the same time, in order to ensure the effective expression of cross modal fusion features, we also added a layer of Transformer block after modal interaction learning to ensure effective modal fusion. Table~\ref{SSM-ViT} reports the effect of the additional Transformer blocks. It is obvious that after adding Transformer blocks, our model has been improved on the original basis, achieving better tracking performance.

\begin{figure}
    \centering
    \includegraphics[width=0.75\linewidth]{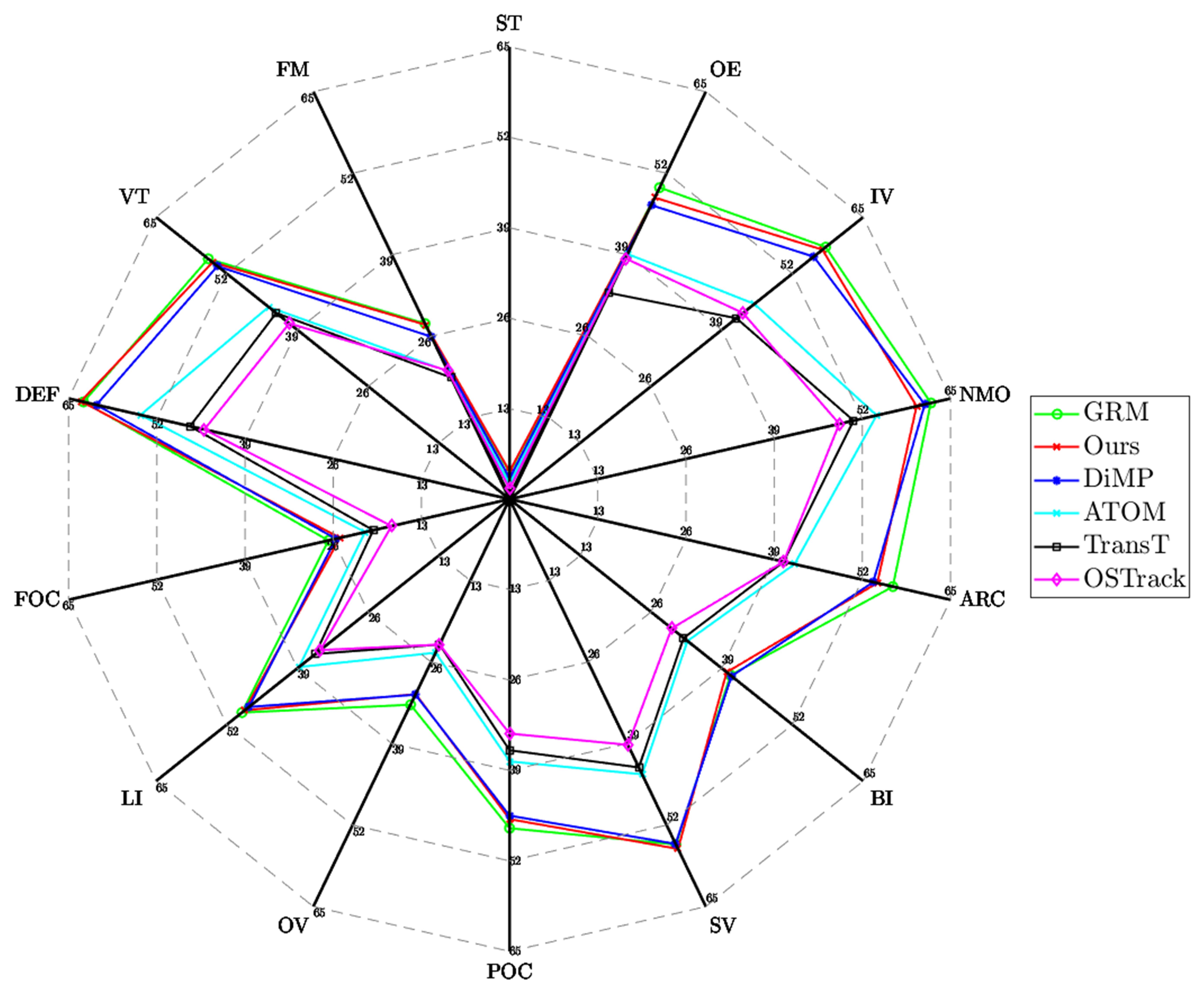}
    \caption{Tracking results under each challenging factor on the FELT SOT dataset.}
    \label{fig:attributeResults}
\end{figure}

\noindent
\textbf{Tracking Results Under Each Attribute.} 
At the same time, we compare our method with other state-of-the-art trackers under each challenging attribute (14 challenging attributes including small target, fast motion, etc.) of FELT. We obtain the results as shown in Fig.~\ref{fig:attributeResults}. It is obvious from the figure that our Mamba-FETrack has reached the most advanced level on attributes, such as SV and DEF, etc. For some other attributes, we have also achieved satisfactory results compared to other current advanced tracking methods. It can be proved that our method can achieve a robust tracking effect in many real-world challenges.

\subsection{Comparison on Tracking Speed, Parameters, and FLOPs}  
In order to verify the low complexity and efficient inference speed of our method, we conduct comparative experiments on three indicators: Tracking Speed, Parameters, and FLOPs. Our method has been compared with mainstream algorithms such as CEUTrack, AMTTrack, and OSTrack. For the one-stream framework OSTrack, we have expanded it to a two-stream framework. The specific approach is to take the Event images and RGB images as inputs and extract the features of the two modalities through the backbone of OSTrack~\cite{ye2022joint}. After that, we concatenate the search features of the two modalities and send them to the tracking head for target object localization. Note that, the OSTrack-B and OSTrack-S are two methods that load ViT-S~\cite{Dosovitskiy2020AnII} and ViT-B~\cite{Dosovitskiy2020AnII} as pre-trained weights, respectively. As shown in Table~\ref{Parameter}, it is evident that the parameters (7MB) of our method have a huge advantage, which is about one-fourteenth of CEUTrack~\cite{tang2022coesot}. Meanwhile, our approach also has lower FLOPs compared to our baseline OSTrack~\cite{ye2022joint}, which only has 59GB. For inference speed, our Mamba-FETrack also achieves 24 FPS. Based on the above comparisons, we can see that our tracker not only has a stable performance during the tracking process but also has a huge advantage in the model's parameters.

\begin{table}
\center
\small     
\caption{Comparison on Tracking Speed, Parameters, and FLOPs.} 
\label{Parameter}
\begin{tabular}{c|ccccc}
\hline  \toprule [0.5 pt]   
\textbf{Tracker}   &\textbf{Ours }  &\textbf{CEUTrack}                 &\textbf{AMTTrack }              &\textbf{OSTrack-B}    &\textbf{OSTrack-S}   \\
\textbf{}          &-               &\cite{tang2022coesot}         &\cite{wang2024longterm}         &\cite{ye2022joint}    &\cite{ye2022joint}                   \\
\hline
\textbf{FPS}            &24           &75         &61            & 63         &84           \\
\textbf{FLOPs (GB)}          &59         &1850      &2070         & 1850      &1076            \\
\textbf{Parameters (MB)}     &7          &97        &108          & 97        &60        \\
\hline \toprule [0.5 pt]   
\end{tabular}
\end{table}

\subsection{Visualization}  
\begin{figure}
    \centering
    \includegraphics[width=1\linewidth]{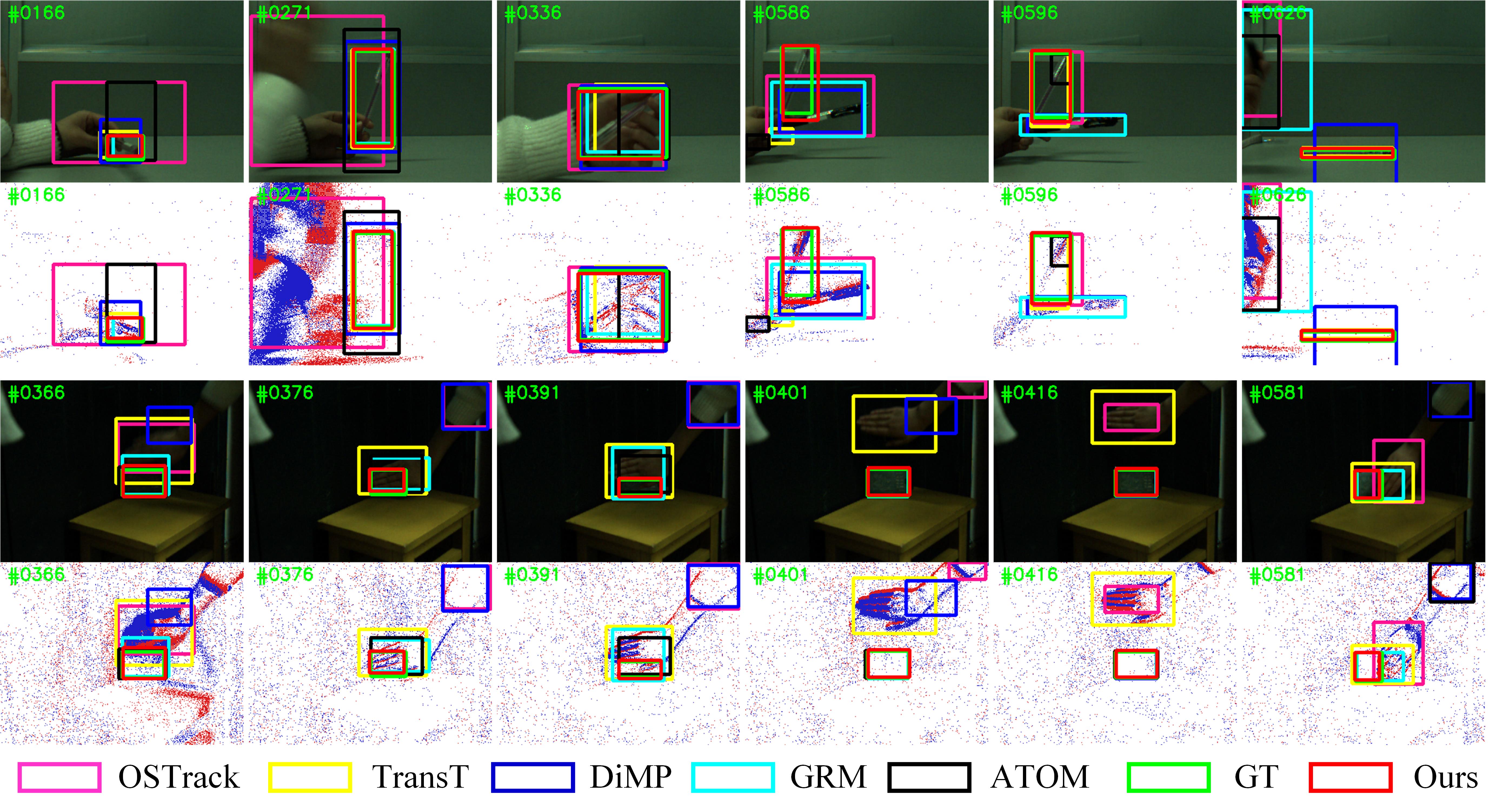}
    \caption{Tracking results of our Mamba-FETrack and other state-of-the-art trackers.} 
    \label{fig:VIS_trackResults}
\end{figure}

\begin{figure}
    \centering
    \includegraphics[width=1\linewidth]{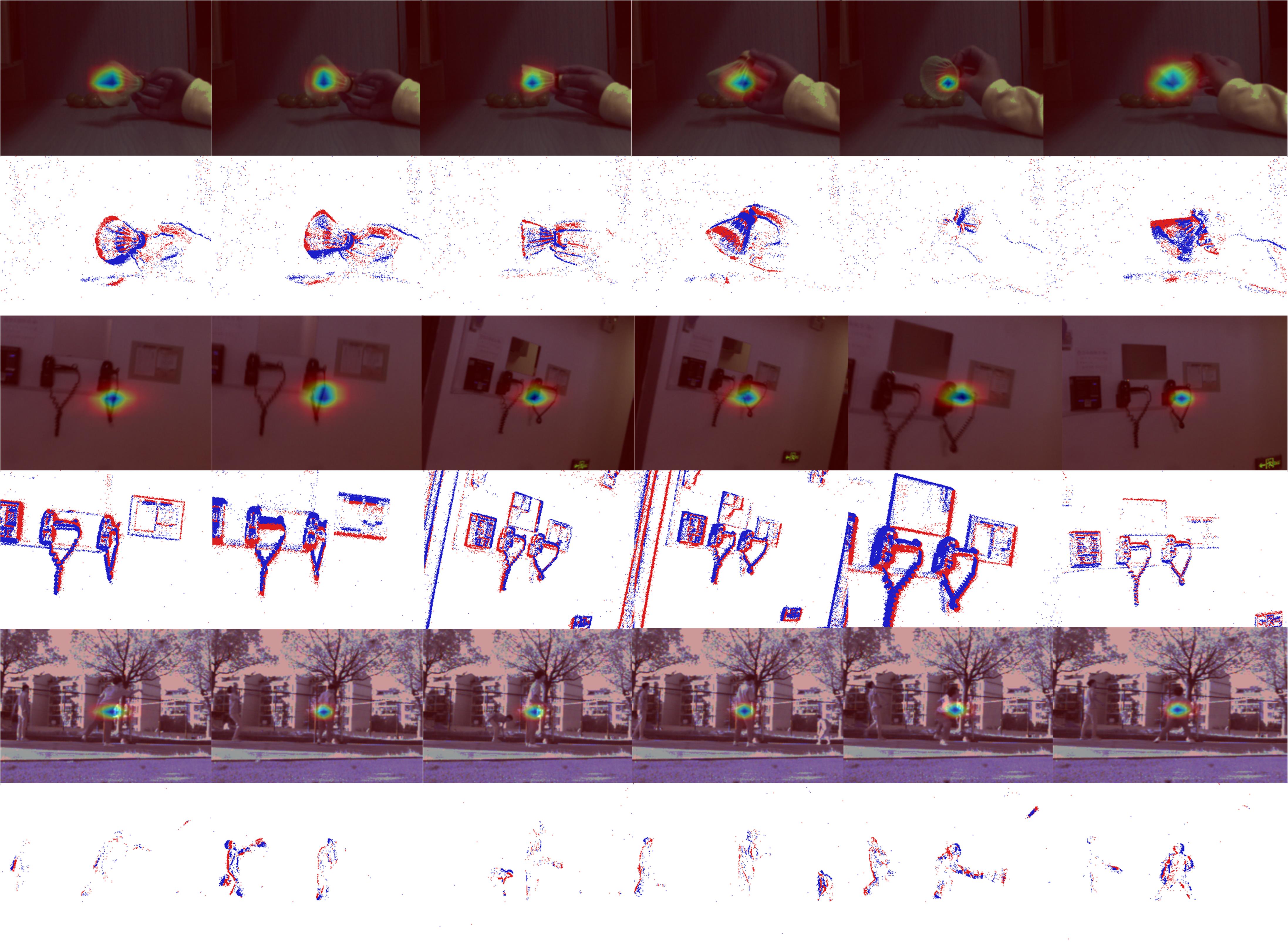}
    \caption{Activation maps predicted by our proposed Mamba-FETrack framework.}
    \label{fig:VIS_activationMAPs}
\end{figure}

In addition to the above quantitative analysis, we also provide some visual displays to help readers more clearly see the advantages of our method. As shown in Fig.~\ref{fig:VIS_trackResults}, which shows the visualization of our method and other trackers in terms of tracking accuracy. It is evident that our Mamba-FETrack is closer to the ground truth bounding boxes compared to other methods. To show the accuracy of our method in target focusing, Fig.~\ref{fig:VIS_activationMAPs} shows the activation maps predicted by our Mamba-FETrack. For different categories, such as the leaves and bicycles in the figure, our method can focus well on the targets. These visualization experiments more intuitively demonstrate the superiority of our method in tracking performance.

\subsection{Limitation Analysis}  
As validated in the experiments, our proposed state space model-based visual tracker achieves a better trade-off between the tracking performance and computational cost on the frame-Event benchmark datasets. It fully demonstrates the promising development prospects of the proposed state space model in the field of multi-modal tracking, particularly in RGB-Event single object tracking. However, our proposed tracker can still be improved from the following aspects: 
1). We simply transform the Event stream into Event images and aggregate them with RGB frames for tracking. Although it is a widely used Event representation, however, it may fail to capture the dynamics of Event points well. 
2). We propose the Mamba-based Frame-Event tracking framework based on existing vision Mamba blocks, but the effective modeling of Event data using Mamba remains an ongoing challenge.

\section{Conclusion and Future Works}    

In this paper, we propose a novel Frame-Event based tracking framework using a state space model, termed Mamba-FETrack. We transform the Event stream into Event images and feed them into the modality-specific encoders along with the RGB frames. Specifically, we adopt the vision Mamba blocks (Vim) to build the backbone networks for the RGB and Event feature learning. Then, we fuse the dual modalities using a FusionMamba block which effectively boosts their interactions. The enhanced features will be fed into a tracking head for target localization. We conducted extensive experiments on two benchmark datasets, which fully validated the effectiveness and efficiency of our tracker for Frame-Event tracking. Through this work, we find that utilizing the Mamba architecture for multi-modal tracking remains an underexplored yet highly promising research direction. Its advantages in terms of memory consumption, computational load, and running speed can be further explored.

In our future works, we will consider improving this framework by conducting more experiments on diverse Event representations, such as Event point, Event voxel, etc. Also, we can design more advanced Mamba architectures for the modeling of Event stream, and RGB-Event fusion procedures. We hope this work can bring some new insights to the tracking field and greatly promote the application of the Mamba architecture in tracking.


{ 
\bibliographystyle{IEEEtran}
\bibliography{reference}

\begin{thebibliography}{10}
\providecommand{\url}[1]{#1}
\csname url@samestyle\endcsname
\providecommand{\newblock}{\relax}
\providecommand{\bibinfo}[2]{#2}
\providecommand{\BIBentrySTDinterwordspacing}{\spaceskip=0pt\relax}
\providecommand{\BIBentryALTinterwordstretchfactor}{4}
\providecommand{\BIBentryALTinterwordspacing}{\spaceskip=\fontdimen2\font plus
\BIBentryALTinterwordstretchfactor\fontdimen3\font minus
  \fontdimen4\font\relax}
\providecommand{\BIBforeignlanguage}[2]{{%
\expandafter\ifx\csname l@#1\endcsname\relax
\typeout{** WARNING: IEEEtran.bst: No hyphenation pattern has been}%
\typeout{** loaded for the language `#1'. Using the pattern for}%
\typeout{** the default language instead.}%
\else
\language=\csname l@#1\endcsname
\fi
#2}}
\providecommand{\BIBdecl}{\relax}
\BIBdecl

\bibitem{Nam2015LearningMC}
H.~Nam and B.~Han, ``Learning multi-domain convolutional neural networks for
  visual tracking,'' \emph{2016 IEEE Conference on Computer Vision and Pattern
  Recognition (CVPR)}, pp. 4293--4302, 2015.

\bibitem{jung2018real}
I.~Jung, J.~Son, M.~Baek, and B.~Han, ``Real-time mdnet,'' in \emph{Proceedings
  of the European conference on computer vision (ECCV)}, 2018, pp. 83--98.

\bibitem{bertinetto2021fullyconvolutional}
L.~Bertinetto, J.~Valmadre, J.~F. Henriques, A.~Vedaldi, and P.~H. Torr,
  ``Fully-convolutional siamese networks for object tracking,'' in
  \emph{Computer Vision--ECCV 2016 Workshops: Amsterdam, The Netherlands,
  October 8-10 and 15-16, 2016, Proceedings, Part II 14}.\hskip 1em plus 0.5em
  minus 0.4em\relax Springer, 2016, pp. 850--865.

\bibitem{xu2020siamfc++}
Y.~Xu, Z.~Wang, Z.~Li, Y.~Yuan, and G.~Yu, ``Siamfc++: Towards robust and
  accurate visual tracking with target estimation guidelines,'' in
  \emph{Proceedings of the AAAI conference on artificial intelligence},
  vol.~34, no.~07, 2020, pp. 12\,549--12\,556.

\bibitem{Tao2016SiameseIS}
R.~Tao, E.~Gavves, and A.~W.~M. Smeulders, ``Siamese instance search for
  tracking,'' \emph{2016 IEEE Conference on Computer Vision and Pattern
  Recognition (CVPR)}, pp. 1420--1429, 2016.

\bibitem{Wang2018SINTRV}
X.~Wang, C.~Li, B.~Luo, and J.~Tang, ``Sint++: Robust visual tracking via
  adversarial positive instance generation,'' \emph{2018 IEEE/CVF Conference on
  Computer Vision and Pattern Recognition}, pp. 4864--4873, 2018.

\bibitem{Chen2021TransformerT}
X.~Chen, B.~Yan, J.~Zhu, D.~Wang, X.~Yang, and H.~Lu, ``Transformer tracking,''
  \emph{2021 IEEE/CVF Conference on Computer Vision and Pattern Recognition
  (CVPR)}, pp. 8122--8131, 2021.

\bibitem{Yan2021LearningST}
B.~Yan, H.~Peng, J.~Fu, D.~Wang, and H.~Lu, ``Learning spatio-temporal
  transformer for visual tracking,'' \emph{2021 IEEE/CVF International
  Conference on Computer Vision (ICCV)}, pp. 10\,428--10\,437, 2021.

\bibitem{Cui2022MixFormerET}
Y.~Cui, J.~Cheng, L.~Wang, and G.~Wu, ``Mixformer: End-to-end tracking with
  iterative mixed attention,'' \emph{2022 IEEE/CVF Conference on Computer
  Vision and Pattern Recognition (CVPR)}, pp. 13\,598--13\,608, 2022.

\bibitem{gao2022aiatrack}
S.~Gao, C.~Zhou, C.~Ma, X.~Wang, and J.~Yuan, ``Aiatrack: Attention in
  attention for transformer visual tracking,'' in \emph{European Conference on
  Computer Vision}.\hskip 1em plus 0.5em minus 0.4em\relax Springer, 2022, pp.
  146--164.

\bibitem{liu2020tanet}
Z.~Liu, X.~Zhao, T.~Huang, R.~Hu, Y.~Zhou, and X.~Bai, ``Tanet: Robust 3d
  object detection from point clouds with triple attention,'' in
  \emph{Proceedings of the AAAI conference on artificial intelligence},
  vol.~34, no.~07, 2020, pp. 11\,677--11\,684.

\bibitem{yang2020interpretable}
D.~Yang, K.~Dyer, and S.~Wang, ``Interpretable deep learning model for online
  multi-touch attribution,'' \emph{arXiv preprint arXiv:2004.00384}, 2020.

\bibitem{huang2020globaltrack}
L.~Huang, X.~Zhao, and K.~Huang, ``Globaltrack: A simple and strong baseline
  for long-term tracking,'' in \emph{Proceedings of the AAAI conference on
  artificial intelligence}, vol.~34, no.~07, 2020, pp. 11\,037--11\,044.

\bibitem{Gallego2019EventBasedVA}
G.~Gallego, T.~Delbr{\"u}ck, G.~Orchard, C.~Bartolozzi, B.~Taba, A.~Censi,
  S.~Leutenegger, A.~J. Davison, J.~Conradt, K.~Daniilidis, and D.~Scaramuzza,
  ``Event-based vision: A survey,'' \emph{IEEE Transactions on Pattern Analysis
  and Machine Intelligence}, vol.~44, pp. 154--180, 2019.

\bibitem{Wang2021VisEventRO}
X.~Wang, J.~Li, L.~Zhu, Z.~Zhang, Z.~Chen, X.~Li, Y.~Wang, Y.~Tian, and F.~Wu,
  ``Visevent: Reliable object tracking via collaboration of frame and event
  flows,'' \emph{IEEE Transactions on Cybernetics}, vol.~54, pp. 1997--2010,
  2021.

\bibitem{tang2022coesot}
C.~Tang, X.~Wang, J.~Huang, B.~Jiang, L.~Zhu, J.~Zhang, Y.~Wang, and Y.~Tian,
  ``Revisiting color-event based tracking: A unified network, dataset, and
  metric,'' \emph{arXiv preprint arXiv:2211.11010}, 2022.

\bibitem{Zhang2022SpikingTF}
J.~Zhang, B.~Dong, H.~Zhang, J.~Ding, F.~Heide, B.~Yin, and X.~Yang, ``Spiking
  transformers for event-based single object tracking,'' \emph{2022 IEEE/CVF
  Conference on Computer Vision and Pattern Recognition (CVPR)}, pp.
  8791--8800, 2022.

\bibitem{wang2024SSMSurvey}
X.~Wang, S.~Wang, Y.~Ding, Y.~Li, W.~Wu, Y.~Rong, W.~Kong, J.~Huang, S.~Li,
  H.~Yang \emph{et~al.}, ``State space model for new-generation network
  alternative to transformers: A survey,'' \emph{arXiv preprint
  arXiv:2404.09516}, 2024.

\bibitem{Nguyen2022S4NDMI}
E.~Nguyen, K.~Goel, A.~Gu, G.~W. Downs, P.~Shah, T.~Dao, S.~A. Baccus, and
  C.~R{\'e}, ``S4nd: Modeling images and videos as multidimensional signals
  using state spaces,'' \emph{ArXiv}, vol. abs/2210.06583, 2022.

\bibitem{Smith2022SimplifiedSS}
J.~Smith, A.~Warrington, and S.~W. Linderman, ``Simplified state space layers
  for sequence modeling,'' \emph{ArXiv}, vol. abs/2208.04933, 2022.

\bibitem{Liu2024VMambaVS}
Y.~Liu, Y.~Tian, Y.~Zhao, H.~Yu, L.~Xie, Y.~Wang, Q.~Ye, and Y.~Liu, ``Vmamba:
  Visual state space model,'' \emph{ArXiv}, vol. abs/2401.10166, 2024.

\bibitem{zhu2024vision}
L.~Zhu, B.~Liao, Q.~Zhang, X.~Wang, W.~Liu, and X.~Wang, ``Vision mamba:
  Efficient visual representation learning with bidirectional state space
  model,'' \emph{arXiv preprint arXiv:2401.09417}, 2024.

\bibitem{Deng2009ImageNetAL}
J.~Deng, W.~Dong, R.~Socher, L.-J. Li, K.~Li, and L.~Fei-Fei, ``Imagenet: A
  large-scale hierarchical image database,'' \emph{2009 IEEE Conference on
  Computer Vision and Pattern Recognition}, pp. 248--255, 2009.

\bibitem{Li2024MambaNDSS}
S.~Li, H.~Singh, and A.~Grover, ``Mamba-nd: Selective state space modeling for
  multi-dimensional data,'' \emph{ArXiv}, vol. abs/2402.05892, 2024.

\bibitem{Xing2024SegMambaLS}
Z.~Xing, T.~Ye, Y.~Yang, G.~Liu, and L.~Zhu, ``Segmamba: Long-range sequential
  modeling mamba for 3d medical image segmentation,'' \emph{ArXiv}, vol.
  abs/2401.13560, 2024.

\bibitem{Ma2024UMambaEL}
J.~Ma, F.~Li, and B.~Wang, ``U-mamba: Enhancing long-range dependency for
  biomedical image segmentation,'' \emph{ArXiv}, vol. abs/2401.04722, 2024.

\bibitem{Ruan2024VMUNetVM}
J.~Ruan and S.~Xiang, ``Vm-unet: Vision mamba unet for medical image
  segmentation,'' \emph{ArXiv}, vol. abs/2402.02491, 2024.

\bibitem{tang2023modeling}
S.~Tang, J.~A. Dunnmon, Q.~Liangqiong, K.~K. Saab, T.~Baykaner, C.~Lee-Messer,
  and D.~L. Rubin, ``Modeling multivariate biosignals with graph neural
  networks and structured state space models,'' in \emph{Conference on Health,
  Inference, and Learning}.\hskip 1em plus 0.5em minus 0.4em\relax PMLR, 2023,
  pp. 50--71.

\bibitem{Wang2024GraphMambaTL}
C.~X. Wang, O.~Tsepa, J.~Ma, and B.~Wang, ``Graph-mamba: Towards long-range
  graph sequence modeling with selective state spaces,'' \emph{ArXiv}, vol.
  abs/2402.00789, 2024.

\bibitem{Behrouz2024GraphMT}
A.~Behrouz and F.~Hashemi, ``Graph mamba: Towards learning on graphs with state
  space models,'' \emph{ArXiv}, vol. abs/2402.08678, 2024.

\bibitem{Liang2024PointMambaAS}
D.~Liang, X.~Zhou, X.~Wang, X.~Zhu, W.~Xu, Z.~Zou, X.~Ye, and X.~Bai,
  ``Pointmamba: A simple state space model for point cloud analysis,''
  \emph{ArXiv}, vol. abs/2402.10739, 2024.

\bibitem{Zhang2024PointCM}
T.~Zhang, X.~Li, H.~Yuan, S.~Ji, and S.~Yan, ``Point cloud mamba: Point cloud
  learning via state space model,'' \emph{ArXiv}, vol. abs/2403.00762, 2024.

\bibitem{Liu2024PointMA}
J.~Liu, R.~Yu, Y.~Wang, Y.~Zheng, T.~Deng, W.~Ye, and H.~Wang, ``Point mamba: A
  novel point cloud backbone based on state space model with octree-based
  ordering strategy,'' \emph{ArXiv}, vol. abs/2403.06467, 2024.

\bibitem{Zubic2024StateSM}
N.~Zubi'c, M.~Gehrig, and D.~Scaramuzza, ``State space models for event
  cameras,'' \emph{ArXiv}, vol. abs/2402.15584, 2024.

\bibitem{islam2022long}
M.~M. Islam and G.~Bertasius, ``Long movie clip classification with state-space
  video models,'' in \emph{European Conference on Computer Vision}.\hskip 1em
  plus 0.5em minus 0.4em\relax Springer, 2022, pp. 87--104.

\bibitem{Wang2023SelectiveSS}
J.~Wang, W.~Zhu, P.~Wang, X.~Yu, L.~Liu, M.~Omar, and R.~Hamid, ``Selective
  structured state-spaces for long-form video understanding,'' \emph{2023
  IEEE/CVF Conference on Computer Vision and Pattern Recognition (CVPR)}, pp.
  6387--6397, 2023.

\bibitem{wang2024longterm}
X.~Wang, J.~Huang, S.~Wang, C.~Tang, B.~Jiang, Y.~Tian, J.~Tang, and B.~Luo,
  ``Long-term frame-event visual tracking: Benchmark dataset and baseline,''
  \emph{arXiv preprint arXiv:2403.05839}, 2024.

\bibitem{Zhang2021ObjectTB}
J.~Zhang, X.~Yang, Y.~Fu, X.~Wei, B.~Yin, and B.~Dong, ``Object tracking by
  jointly exploiting frame and event domain,'' \emph{2021 IEEE/CVF
  International Conference on Computer Vision (ICCV)}, pp. 13\,023--13\,032,
  2021.

\bibitem{Zhang2023FrameEventAA}
J.~Zhang, Y.~Wang, W.~Liu, M.~Li, J.~Bai, B.~Yin, and X.~Yang, ``Frame-event
  alignment and fusion network for high frame rate tracking,'' \emph{2023
  IEEE/CVF Conference on Computer Vision and Pattern Recognition (CVPR)}, pp.
  9781--9790, 2023.

\bibitem{Zhu2023CrossmodalOH}
Z.~Zhu, J.~Hou, and D.~O. Wu, ``Cross-modal orthogonal high-rank augmentation
  for rgb-event transformer-trackers,'' \emph{2023 IEEE/CVF International
  Conference on Computer Vision (ICCV)}, pp. 21\,988--21\,998, 2023.

\bibitem{Zheng2024ODTrackOD}
Y.~Zheng, B.~Zhong, Q.~Liang, Z.~Mo, S.~Zhang, and X.~Li, ``Odtrack: Online
  dense temporal token learning for visual tracking,'' \emph{ArXiv}, vol.
  abs/2401.01686, 2024.

\bibitem{Gu2021CombiningRC}
A.~Gu, I.~Johnson, K.~Goel, K.~K. Saab, T.~Dao, A.~Rudra, and C.~R'e,
  ``Combining recurrent, convolutional, and continuous-time models with linear
  state-space layers,'' in \emph{Neural Information Processing Systems}, 2021.

\bibitem{gu2020hippo}
A.~Gu, T.~Dao, S.~Ermon, A.~Rudra, and C.~R{\'e}, ``Hippo: Recurrent memory
  with optimal polynomial projections,'' \emph{Advances in neural information
  processing systems}, vol.~33, pp. 1474--1487, 2020.

\bibitem{gu2021efficiently}
A.~Gu, K.~Goel, and C.~R{\'e}, ``Efficiently modeling long sequences with
  structured state spaces,'' \emph{arXiv preprint arXiv:2111.00396}, 2021.

\bibitem{Gu2023MambaLS}
A.~Gu and T.~Dao, ``Mamba: Linear-time sequence modeling with selective state
  spaces,'' \emph{ArXiv}, vol. abs/2312.00752, 2023.

\bibitem{He2024PanMambaEP}
X.~He, K.~Cao, K.~R. Yan, R.~Li, C.~Xie, J.~Zhang, and M.~Zhou, ``Pan-mamba:
  Effective pan-sharpening with state space model,'' \emph{ArXiv}, vol.
  abs/2402.12192, 2024.

\bibitem{kalman1960new}
R.~E. Kalman, ``A new approach to linear filtering and prediction problems,''
  \emph{Journal of Basic Engineering}, vol.~82, pp. 35--45, 1960.

\bibitem{ye2022joint}
B.~Ye, H.~Chang, B.~Ma, S.~Shan, and X.~Chen, ``Joint feature learning and
  relation modeling for tracking: A one-stream framework,'' in \emph{European
  conference on computer vision}.\hskip 1em plus 0.5em minus 0.4em\relax
  Springer, 2022, pp. 341--357.

\bibitem{loshchilov2018adamw}
I.~Loshchilov and F.~Hutter, ``Decoupled weight decay regularization,'' in
  \emph{International Conference on Learning Representations}, 2018.

\bibitem{paszke2019pytorch}
A.~Paszke, S.~Gross, F.~Massa, A.~Lerer, J.~Bradbury, G.~Chanan, T.~Killeen,
  Z.~Lin, N.~Gimelshein, L.~Antiga \emph{et~al.}, ``Pytorch: An imperative
  style, high-performance deep learning library,'' \emph{Advances in neural
  information processing systems}, vol.~32, 2019.

\bibitem{Li2018HighPV}
B.~Li, J.~Yan, W.~Wu, Z.~Zhu, and X.~Hu, ``High performance visual tracking
  with siamese region proposal network,'' \emph{2018 IEEE/CVF Conference on
  Computer Vision and Pattern Recognition}, pp. 8971--8980, 2018.

\bibitem{Chen2020SiameseBA}
Z.~Chen, B.~Zhong, G.~Li, S.~Zhang, and R.~Ji, ``Siamese box adaptive network
  for visual tracking,'' \emph{2020 IEEE/CVF Conference on Computer Vision and
  Pattern Recognition (CVPR)}, pp. 6667--6676, 2020.

\bibitem{bhat2020know}
G.~Bhat, M.~Danelljan, L.~Van~Gool, and R.~Timofte, ``Know your surroundings:
  Exploiting scene information for object tracking,'' in \emph{Computer
  Vision--ECCV 2020: 16th European Conference, Glasgow, UK, August 23--28,
  2020, Proceedings, Part XXIII 16}.\hskip 1em plus 0.5em minus 0.4em\relax
  Springer, 2020, pp. 205--221.

\bibitem{dong2020clnet}
X.~Dong, J.~Shen, L.~Shao, and F.~Porikli, ``Clnet: A compact latent network
  for fast adjusting siamese trackers,'' in \emph{European Conference on
  Computer Vision}.\hskip 1em plus 0.5em minus 0.4em\relax Springer, 2020, pp.
  378--395.

\bibitem{Danelljan2018ATOMAT}
M.~Danelljan, G.~Bhat, F.~S. Khan, and M.~Felsberg, ``Atom: Accurate tracking
  by overlap maximization,'' \emph{2019 IEEE/CVF Conference on Computer Vision
  and Pattern Recognition (CVPR)}, pp. 4655--4664, 2018.

\bibitem{Bhat2019LearningDM}
G.~Bhat, M.~Danelljan, L.~V. Gool, and R.~Timofte, ``Learning discriminative
  model prediction for tracking,'' \emph{2019 IEEE/CVF International Conference
  on Computer Vision (ICCV)}, pp. 6181--6190, 2019.

\bibitem{Danelljan2020ProbabilisticRF}
M.~Danelljan, L.~V. Gool, and R.~Timofte, ``Probabilistic regression for visual
  tracking,'' \emph{2020 IEEE/CVF Conference on Computer Vision and Pattern
  Recognition (CVPR)}, pp. 7181--7190, 2020.

\bibitem{Gao2023GeneralizedRM}
S.~Gao, C.~Zhou, and J.~Zhang, ``Generalized relation modeling for transformer
  tracking,'' \emph{2023 IEEE/CVF Conference on Computer Vision and Pattern
  Recognition (CVPR)}, pp. 18\,686--18\,695, 2023.

\bibitem{Mayer2022TransformingMP}
C.~Mayer, M.~Danelljan, G.~Bhat, M.~Paul, D.~P. Paudel, F.~Yu, and L.~V. Gool,
  ``Transforming model prediction for tracking,'' \emph{2022 IEEE/CVF
  Conference on Computer Vision and Pattern Recognition (CVPR)}, pp.
  8721--8730, 2022.

\bibitem{Dosovitskiy2020AnII}
A.~Dosovitskiy, L.~Beyer, A.~Kolesnikov, D.~Weissenborn, X.~Zhai,
  T.~Unterthiner, M.~Dehghani, M.~Minderer, G.~Heigold, S.~Gelly, J.~Uszkoreit,
  and N.~Houlsby, ``An image is worth 16x16 words: Transformers for image
  recognition at scale,'' \emph{ArXiv}, vol. abs/2010.11929, 2020.

\end{thebibliography}
}
\end{document}